%% file: gradient_free.tex
\documentclass{article}
\usepackage{ijcai23}

\usepackage{times}
\usepackage{soul}
\usepackage{url}
\usepackage{bbm}
\usepackage{mathrsfs}
\usepackage{subcaption}
\usepackage[hidelinks]{hyperref}
\usepackage[utf8]{inputenc}
\usepackage{caption}
\usepackage{graphicx}
\usepackage{amsmath, bm}
\usepackage{amsthm}
\usepackage{booktabs}
\usepackage{algorithm}
\usepackage{algorithmic}
\usepackage[switch]{lineno}
\usepackage{xcolor}
\usepackage{bbm}
\usepackage[justification=centering]{caption}

\DeclareMathOperator{\Tr}{Tr}

\linenumbers

\newtheorem{theorem}{Theorem}

\title{Simple yet Effective Gradient-Free Graph Convolutional Networks}

\author{
Yulin Zhu$^1$
\and
Xing Ai$^1$\and
Qimai Li$^1$\and
Xiao-Ming Wu$^{1}$\and
Kai Zhou$^1$
\affiliations
$^1$The Hong Kong Polytechnic University
\emails
\{yulinzhu, xiao-ming.wu, kaizhou\}@polyu.edu.hk,
\{xing96.ai, csqmli\}@connect.polyu.hk
}

\begin{document}
\nolinenumbers
\maketitle

\begin{abstract}
    Linearized Graph Neural Networks (GNNs) have attracted great attention in recent years for graph representation learning. Compared with nonlinear Graph Neural Network (GNN) models, linearized GNNs are much more time-efficient and can achieve comparable performances on typical downstream tasks such as node classification. Although some linearized GNN variants are purposely crafted to mitigate ``over-smoothing", empirical studies demonstrate that they still somehow suffer from this issue. In this paper, we instead relate over-smoothing with the vanishing gradient phenomenon and craft a gradient-free training framework to achieve more efficient and effective linearized GNNs which can significantly overcome over-smoothing and enhance the generalization of the model. The experimental results demonstrate that our methods achieve better and more stable performances on node classification tasks with varying depths and cost much less training time.
\end{abstract}

\input{sections/intro}

\input{sections/background}

\input{sections/method}
\input{sections/exp}
\input{sections/conclusion}
\clearpage

\bibliographystyle{named}
\bibliography{gradient_free}

\end{document}

%% file: sections/intro.tex
\section{Introduction}
Graph Neural Networks (GNNs) are playing an increasingly important role in mining relational data represented as graphs. They achieved state-of-the-art performances on various tasks, including 
node classification \cite{GCN,GAT},
link prediction \cite{linkpred},
graph classification \cite{powerful},
and node clustering \cite{graphclustering}.
A particularly successful type of GNNs is Graph Convolutional Networks (GCNs) \cite{GCN}. 
Recently, it is argued that unlike other deep learning models such as Convolutional Neural Networks (CNNs) \cite{CNN} in the computer vision domain, the effectiveness of GCNs does not come from the non-linear operations (e.g., the non-linear activation functions) but is a result of the nature of the low pass filter that can locally smooth the node features over the graph. 
Specifically, \citeauthor{SGC} crafted a simplified GCN by iteratively removing the non-linear activations and degenerating the multi-layer GCN to logistic regression with one weight matrix. This simplified model is termed  Simplified Graph Convolutional Network (SGC). Empirically, it was  shown that SGC could achieve comparable performances as other nonlinear competitors in most cases. But, SGC enjoys much less training time since training SGC is a simple multiclass logistic regression problem.
The benefits of SGC are not limited to computational savings. Recently in the security domain, SGC is commonly used as a surrogate model to simplify the attack design~\cite{nettack,mettack,binarizedattack}. In addition, SGC could potentially reduce the computational complexity for secure computation schemes~\cite{mohassel2018aby3,rathee2020cryptflow2}, for which dealing with nonlinear operations are notoriously expensive.

Thus, following SGC, a lot of research efforts have been devoted to studying and further improving such linearized GCNs. Notably, \citeauthor{SSGC,DGC} have shown that SGC would suffer from the over-smoothing problem~\cite{deepinsight} if more graph convolutional layers are stacked, which could cause a severe downgrade in the node classification performance. The above works tackle the over-smoothing problem from the perspective of the graph spectrum. More specifically, they design a suitable graph filter to balance the local and global information obtained from the propagation step to prevent the node from paying unnecessarily more attention to its neighbors in the receptive field and ignoring self-information. In this paper, we instead relate the over-smoothing issue to the ineffective gradient-descent-based training process, which further results in the downgrade in node classification performance.
Specifically, it has been shown that stacking more layers to CNN or RNN \cite{RNN} will cause the \emph{vanishing gradient} problem~\cite{vanishgradient}, meaning that the gradients used to update the neural network's weights will be minuscule during the training phase and lead to insufficient training of the deep neural networks. 
Unfortunately, we observe that the linearized GCNs also have this problem (see Fig.~\ref{fig-vanishing-gradient}). 

To address this issue, we propose a \textbf{gradient-free} solution for the training of a series of linearized GCN models. In short, our method does not require training the GCN model using gradient descent (thus gradient-free). Instead, we directly calculate the optimal model parameters through closed-form functions. Specifically, we replace and kernelize the original negative log-likelihood (NLL) loss of the linearized GCNs to a dual formulation and map the graph filters obtained from the linearized GCNs to a reproducing kernel Hilbert space (RKHS) to train the node classifier. We transform the primal classification problem into a regression problem on the RKHS. By setting the first derivative
of the dual form loss with respect to the Lagrange multiplier to zero, we can obtain a closed-form solution for the model parameters with a nice \textit{scale invariance} property.
That is, we can directly obtain the optimal weights for linearized GCNs, avoiding the tedious gradient-descent-based training and tuning various hyperparameters (e.g., learning rate, epoch number, momentum, etc.).
Importantly, since our method is gradient-free, it can directly prevent the vanishing gradient problem when the number of layers $K$ is large. 

Overall, our methods can serve as a framework for refining linearized GCNs. We use three representative linearized GCN models (i.e., SGC, SSGC, and DGC) as examples, for which our methods lead to improved models termed gfSGC, gfSSGC, and gfDGC, respectively.
The experimental results show that our proposed gradient-free linearized GCNs outperforms other baseline methods regardless of the number of GCN layers. 
In addition, we also empirically show that our methods have better performances in terms of computational efficiency and stability.

\subsection{Related Works}
Graph convolutional networks~\cite{GCN} are powerful tools to mine non-Euclidean relational data. There are two typical analytical methods for GCNs: spectrum-based and spatial-based. Spectrum-based GCNs~\cite{GCN,revisitgcn} design the graph layer via localized first-order approximation of spectral graph convolutions in the Fourier domain. Spatial-based GCNs~\cite{GraphSage,SpatialGCN} instead propagate and update the node features based on the topology relationship between the target node and its neighbors. 
Linearized GCNs simplify the complex nonlinear GCNs to the multi-class logistic regressions (SGC~\cite{SGC}), which achieve comparable node classification performances to vanilla GCNs. On the other hand, Simple Spectral Graph Convolution (SSGC)~\cite{SSGC} and Decoupled Graph Convolution (DGC)~\cite{DGC} tackle the over-smoothing problem encountered by SGC via crafted tricks to prevent the node features from paying too much attention to the neighbors' information when the depth $K$ increases.


%% file: sections/background.tex
\section{Preliminaries}

\subsection{Notations}
We denote $\mathcal{G}=\{\mathbf{A},\mathbf{X}\}$ as a unweighted attributed graph, where $\mathbf{A}\in\mathbbm{R}^{N\times N}$ is the adjacency matrix with binary entries and $\mathbf{X}$ is the attribute matrix, $N$ is the node number. In this paper, we focus on the semi-supervised node classification task \cite{GCN}. Under this scenario, each node is assigned to a label $y_{i}$ which represents a class $c_{i}$. We denote $\mathbf{Y}=\{y_{i}\}_{i=1}^{N}$ and $C$ as the label set and class number. The goal of the node classification task is to achieve a good performance on predicting the nodes' labels in the testing set.

\subsection{Representative Linearized GCNs}
\subsubsection{Simplified Graph Convolution (SGC)}
SGC is the first linearized GCN model for graph data. Intuitively, \citeauthor{SGC} argued that the powerful representation learning of the GCNs does not come from the non-linear activations like $\text{ReLU}(\cdot)$. To address this issue, it degenerates the vanilla GCN by repeatedly erasing the nonlinear activations of each graph convolutional layer and then using a single weight matrix to represent the stacked weight matrices for each layer. As a result, SGC can be formulated as:
\begin{subequations}
    \begin{align}
    \label{sgc}
    &\widehat{\mathbf{Y}}=\text{softmax}(\hat{\mathbf{A}}^{K}\mathbf{X}\mathbf{W}), \\ 
    \label{sgc-laplacian-normalized-matrix}
    &\hat{\mathbf{A}}=\Tilde{\mathbf{D}}^{-\frac{1}{2}}\Tilde{\mathbf{A}}\Tilde{\mathbf{D}}^{-\frac{1}{2}}, \ \Tilde{\mathbf{A}}=\mathbf{A}+\mathbf{I}.
    \end{align}
\end{subequations}
Eqn.~\eqref{sgc} can be regarded as a multi-class logistic regression \cite{multiclassLR} with the complex nodal features $\hat{\mathbf{A}}^{K}\mathbf{X}$ as the inputs, where $K$ is the number of layers. Then, we can implement Adam optimizer \cite{adam} to train the multi-class logistic regression.

\subsubsection{Simple Spectral Graph Convolution (SSGC)}
To tackle the over-smoothing issue, SSGC goes a step further to augment the modified Markov diffusion kernel \cite{MDK} with SGC which can trade off between low and high pass filter bands in the spectrum domain, which can significantly alleviate the over-smoothing phenomenon. Then, the SSGC is formulated as:
\begin{equation}
    \label{ssgc}
    \widehat{\mathbf{Y}}=\text{softmax}(\frac{1}{K}\sum_{k=1}^{K}((1-\tau)\hat{\mathbf{A}}^{k}\mathbf{X}+\tau\mathbf{X})\mathbf{W}), 
\end{equation}
where $\hat{\mathbf{A}}$ is the normalized Laplacian matrix defined in Eqn.~\eqref{sgc-laplacian-normalized-matrix} and $\tau$ is a hyperparameter to trade off between the self-information and consecutive neighbors' information. Similarly, we can regard the complex nodal features $\frac{1}{K}\sum_{k=1}^{K}((1-\tau)\hat{\mathbf{A}}^{k}\mathbf{X}+\tau\mathbf{X})$ as the inputs to the multi-class logistic regression for the downstream semi-supervised node classification task. 

\subsubsection{Decoupled Graph Convolution (DGC)}
DGC regards the propagation of SGC as a numerical discretization of the graph heat equation~\cite{GHE}. \citeauthor{DGC} utilize the Euler method with terminal time to update the features to approximate the graph heat equation with $K$ forward steps. Then, \citeauthor{DGC} decouples the terminal time $T$ and the depth $K$ to get a more fine-grained approximation to the vanilla propagation steps to mitigate the over-smoothing. It crafts the propagation step for each layer as:
\begin{subequations}
    \label{dgc}
    \begin{align}
        \label{eqn-dgc-propagation}
        \widehat{\mathbf{Y}}=softmax((1-\frac{T}{K})\mathbf{I}+\frac{T}{K}\hat{\mathbf{A}})^{K}\mathbf{X}
    \end{align}
\end{subequations}
where $T$ is the terminal time to balance the under-smoothing and over-smoothing. Then, the hybrid nodal features are fed into a logistic regression for training.

\section{Vanishing Gradient and Over-smoothing}
\begin{figure}[!th]
	\centering
	\begin{subfigure}[b]{0.234\textwidth}
		\centering
		\includegraphics[width=\textwidth,height=2.cm]{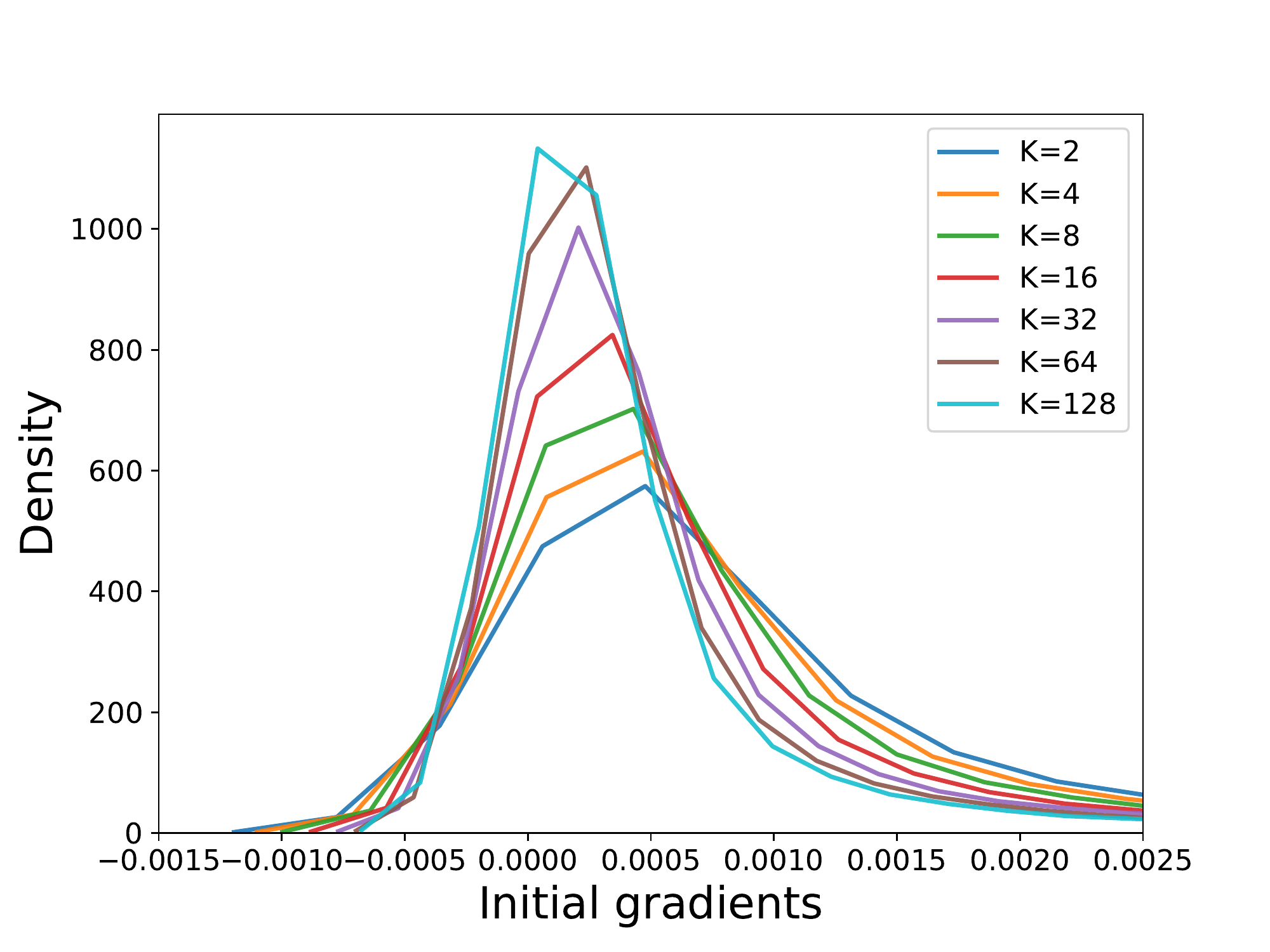}
		\caption{SGC initial gradients}
        \label{fig-SGC-initial-gradients}
	\end{subfigure}
 	\hfill
	\begin{subfigure}[b]{0.234\textwidth}
		\centering
		\includegraphics[width=\textwidth,height=2.cm]{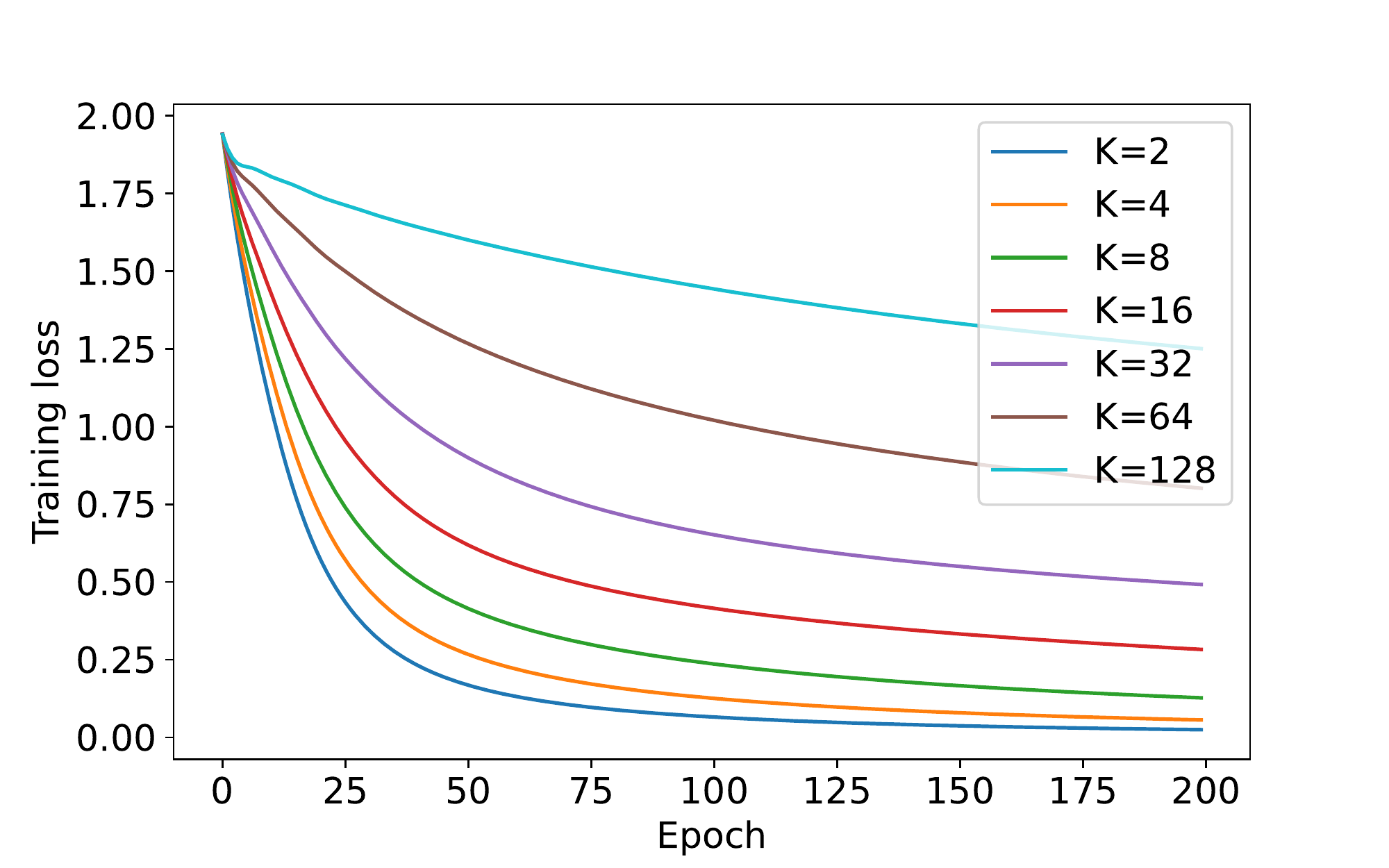}
		\caption{SGC loss trace plot}
        \label{fig-SGC-loss-trace}
	\end{subfigure}
 \hfill
	\begin{subfigure}[b]{0.234\textwidth}
		\centering
		\includegraphics[width=\textwidth,height=2.cm]{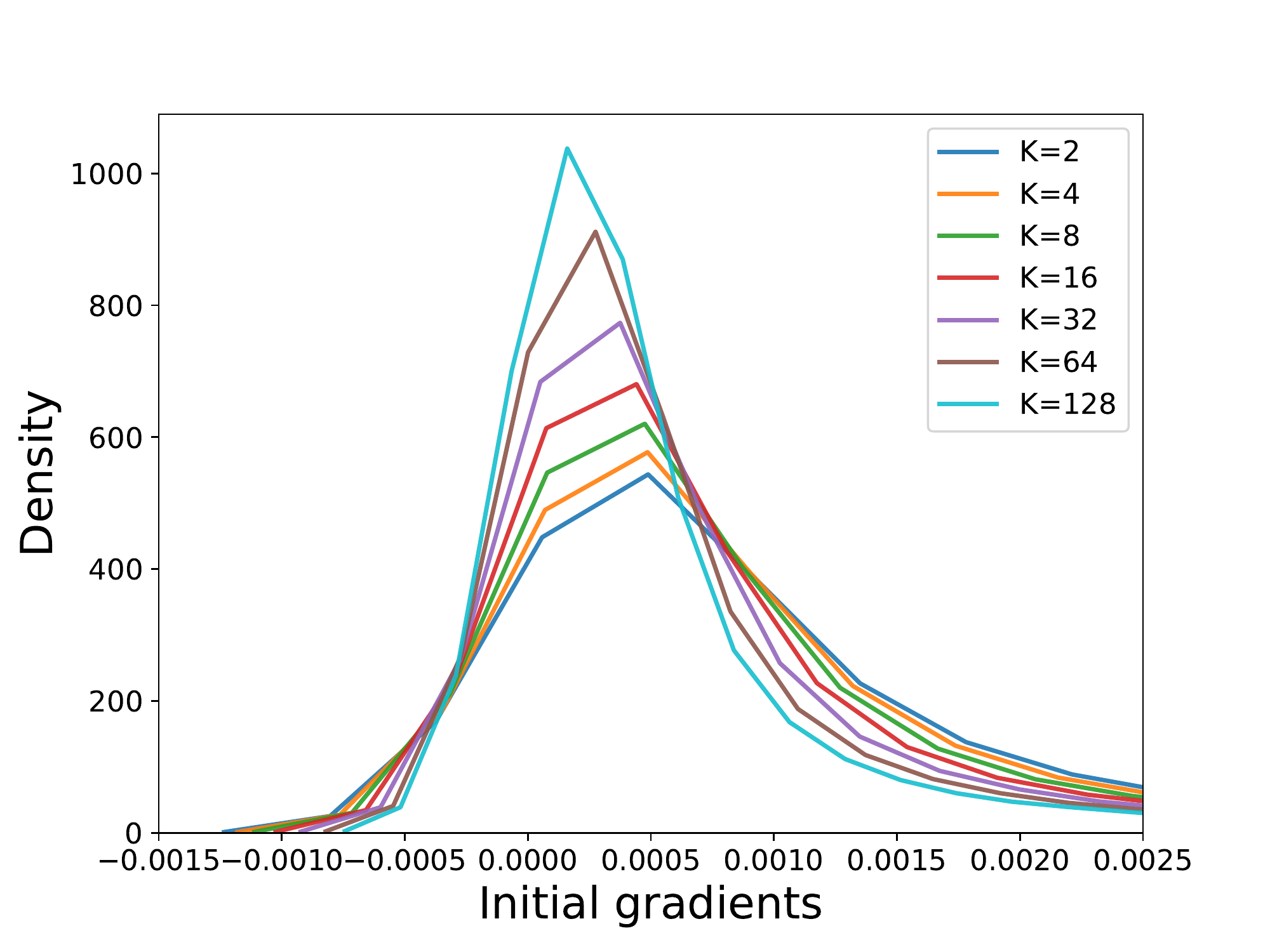}
		\caption{SSGC initial gradients}
        \label{fig-SSGC-initial-gradients}
	\end{subfigure}
 	\hfill
	\begin{subfigure}[b]{0.234\textwidth}
		\centering
		\includegraphics[width=\textwidth,height=2.cm]{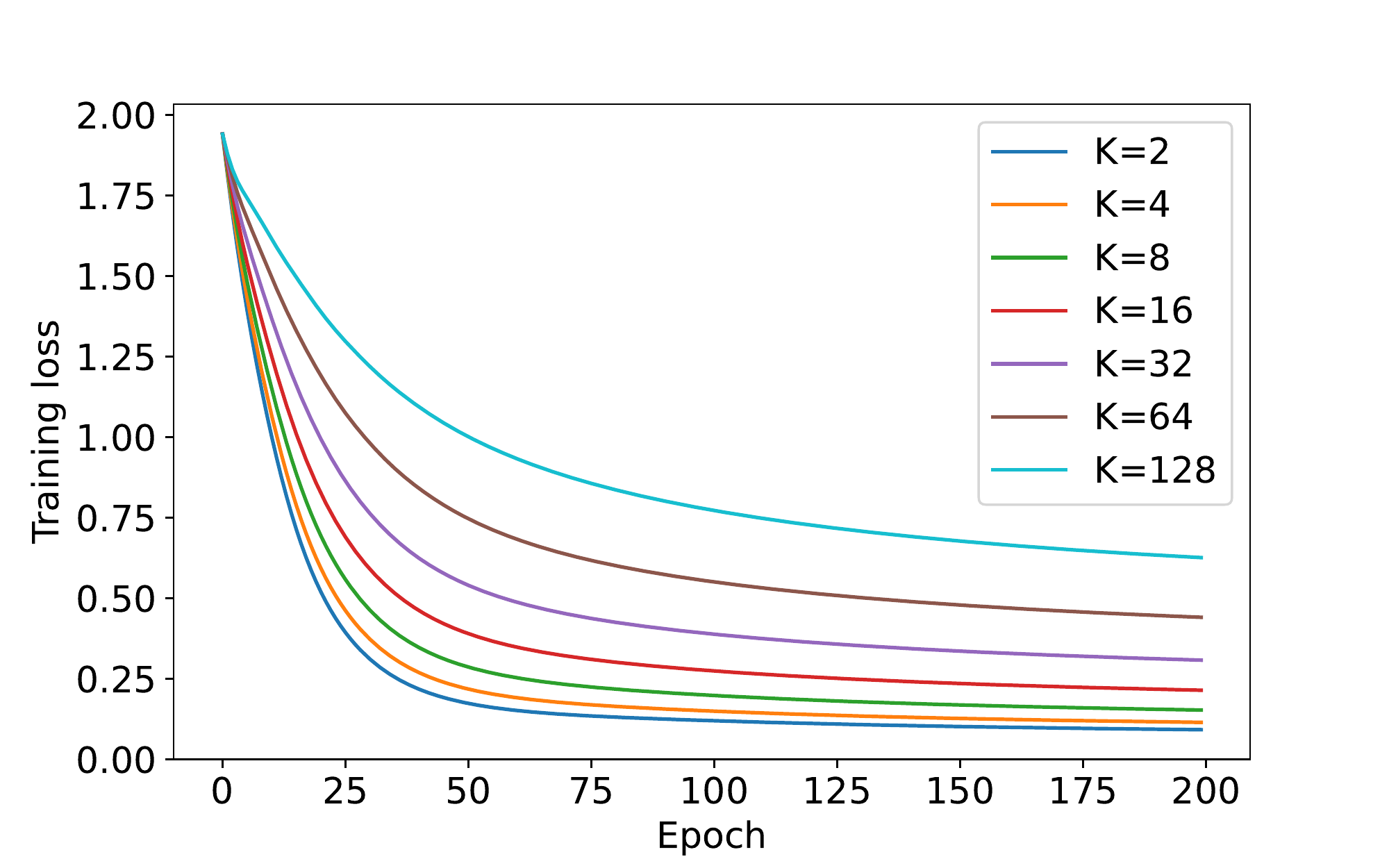}
		\caption{SSGC loss trace plot}
        \label{fig-SSGC-loss-trace}
	\end{subfigure}
    \hfill
	\begin{subfigure}[b]{0.234\textwidth}
		\centering
		\includegraphics[width=\textwidth,height=2.cm]{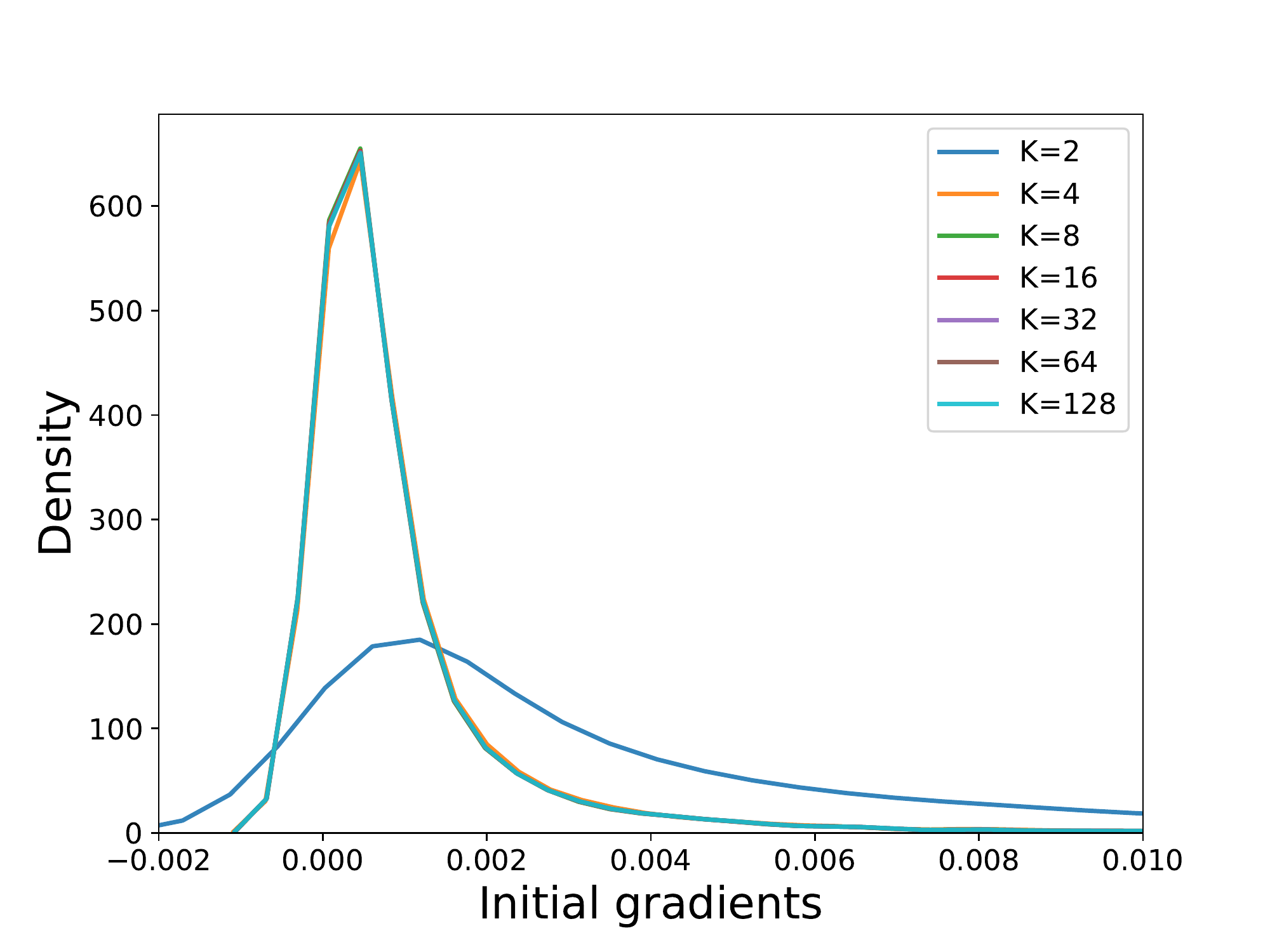}
		\caption{DGC initial gradients}
        \label{fig-DGC-initial-gradients}
	\end{subfigure}
     \hfill
    	\begin{subfigure}[b]{0.234\textwidth}
		\centering
		\includegraphics[width=\textwidth,height=2.cm]{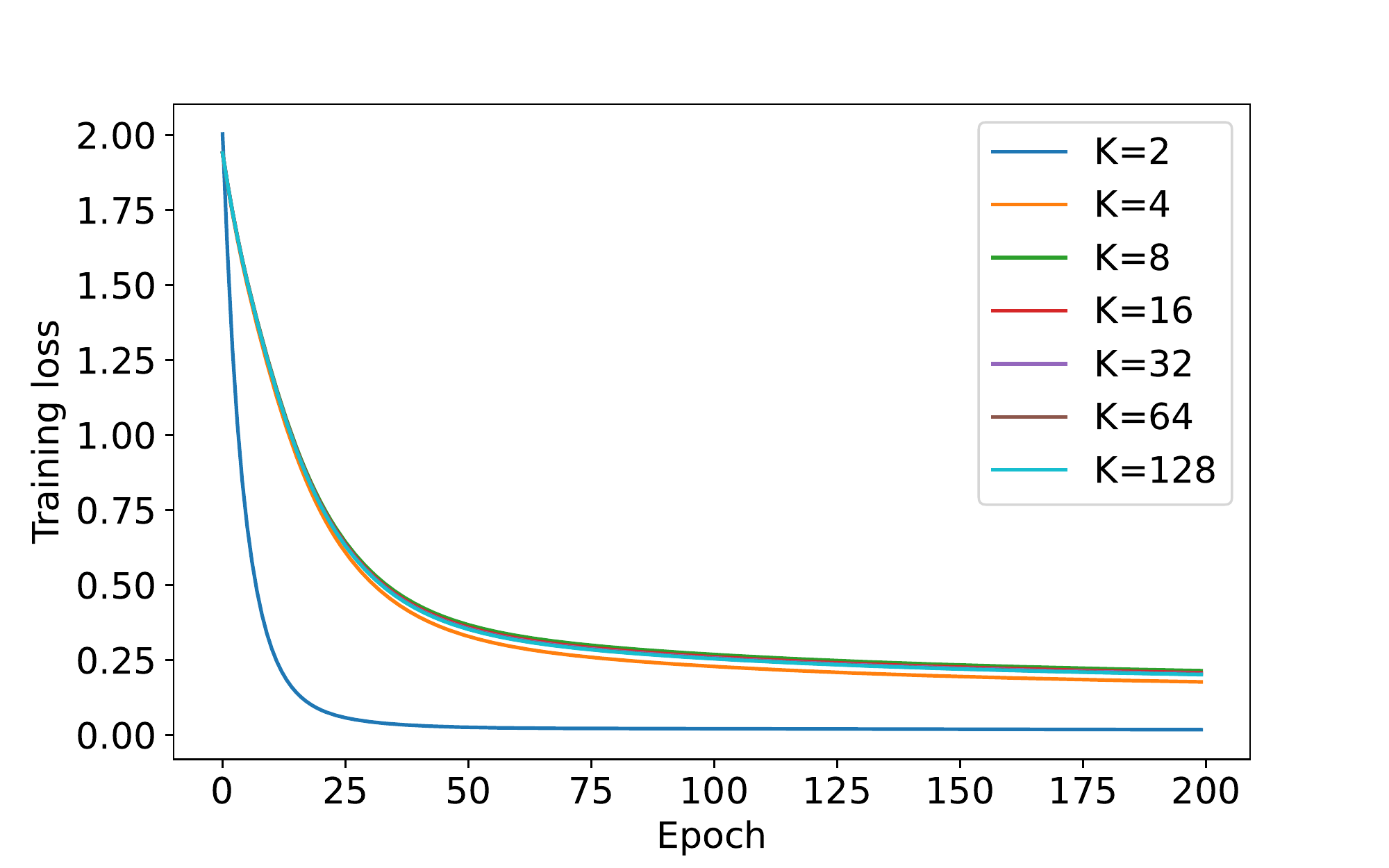}
		\caption{DGC loss trace plot}
        \label{fig-DGC-loss-trace}
	\end{subfigure}
	\caption{Gradient density plot and the trace plot of training loss for SGC, SSGC and DGC. SGC and SSGC encounter the vanishing gradient problem while DGC with relatively small $T$ can mitigate this issue.}
	\label{fig-vanishing-gradient}
\end{figure}

Linearized GCNs suffer from the over-smoothing issue when $K$ increases. A possible cause is that the vanishing gradient problem prevents the model from being effectively trained through gradient descent. In this section, we provide both theoretical and empirical analysis (Fig.~\ref{fig-vanishing-gradient}) of the vanishing gradient problem of linearized GCNs.

For convenience, we use $\mathcal{F}_{G}$ (Eqn.~\eqref{fextra}) to represent the graph filters obtained from the linearized GCNs.
Recall that the feed-forward function 
of linearized GCN $\widehat{\mathbf{Y}}=\text{softmax}(\mathcal{F}_{G}\mathbf{W})$
is trained by minimizing cross-entropy loss: $\min_{\mathbf{W}} \mathcal{L}_{CE}{(\mathbf{Y}, \widehat{\mathbf{Y}})}$.
By the chain rule, we have
\begin{equation}
    \frac{\partial \mathcal{L}_{CE}}{\partial \mathbf{W}} = \mathcal{F}_{G}^\top \frac{\partial \mathcal{L}_{CE}}{\partial \mathbf{H}}, 
\end{equation}
where $\mathbf{H}\overset{\Delta}{=}\mathcal{F}_{G}\mathbf{W}$. The gradient of the model parameters $\mathbf{W}$ is significantly affected by the scale of $\mathcal{F}_{G}$. The parameters $\mathbf{W}$ suffer from the vanishing gradient problem if $\mathcal{F}_{G}$ is too small, which is true when $K$ is large. Since the existence of bias parameters, logistic regression is invariant to data point translation. Thus, we will focus on our discussion on the scale of zero-centered $\mathcal{F}_{G}$ defined as follows:
\begin{equation}
    \mathcal{F}_G - \frac1N\mathbf{1}_N\mathbf{1}_N^\top\mathcal{F}_{G},
\end{equation}
where $\mathbf{1}_N\in\mathbbm{R}^{N}$ is an all-one vector, and $\frac1N\mathbf{1}_N^\top\mathcal{F}_{G}$ is the center of data points. Theorem~\ref{thm-vanish-grad} states the conditions for the vanishment of the scale of the zero-centered $\mathcal{F}_{G}$ as $K$ increases.

\begin{theorem}
\label{thm-vanish-grad}
For SGC, zero-centered $\mathcal{F}_{G}$ approximates a zero vector as  $K$ approximates $+\infty$,
if the graph is connected and not bipartite. That is,
\begin{equation}\label{eq:sgc_dgc_lim}
\lim_{K\to\infty} (\mathcal{F}_{G} - \frac1N\mathbf{1}_N\mathbf{1}_N^\top\mathcal{F}_{G}) = \mathbf{0}.
\end{equation}
For SSGC, it approximates zero-centered $\tau \mathbf{X}$, which is small if $\tau$ is small. That is,
\begin{equation}\label{eq:ssgc_lim}
\resizebox{0.91\linewidth}{!}{$
\lim_{K\to\infty} \left(\mathcal{F}_{G} - \frac1N\mathbf{1}_N\mathbf{1}_N^\top\mathcal{F}_{G}\right) =
    \tau \left(\mathbf{X} - \frac1N\mathbf{1}_N\mathbf{1}_N^\top\mathbf{X}\right).
$}
\end{equation}
For DGC, let $\mathbf{A}_{rw}$ denote the row-normalized adjacency matrix. We have
\begin{align}\label{eq:dgc_lim}
\lim_{K\to\infty} & \left(\mathcal{F}_{G} - \tfrac1N\mathbf{1}_N\mathbf{1}_N^\top\mathcal{F}_{G}\right) \\ 
    &= (I - \tfrac1N\mathbf{1}_N\mathbf{1}_N^\top)\exp(T(\mathbf{A}_{rw}-I))\mathbf{X},
\end{align}
which is small, when $T$ is large.
\end{theorem}
The proof of Theorem~\ref{thm-vanish-grad} is presented in Sec.~C of the supplement.
We note that the analysis in the proof applies to row-normalized adjacency matrix $\mathbf{A}_{rw}$, while the most commonly adopted one is symmetrically normalized $\hat{\mathbf{A}}$. However, for $\hat{\mathbf{A}}$
the left side of Eqn. \eqref{eq:sgc_dgc_lim} converges to a small vector instead of zero. As a result, the vanishing gradient problem still exists.

We also demonstrate the vanishing gradient problem empirically. Fig.~\ref{fig-SGC-initial-gradients}, \ref{fig-SSGC-initial-gradients} and \ref{fig-DGC-initial-gradients} present the density of the gradients with respect to the model parameters at the initial iteration for the three linearized GCN models, respectively. (We obtain the gradients of $\mathbf{W}$ at first iteration and reshape it to one vector and curve the distribution of the vector with density plot.)
It is observed that the gradients of SGC and SSGC will indeed vanish as we increase the number of layers $K$. 
Meanwhile, we also trace the training losses (Fig.~\ref{fig-SGC-loss-trace}, \ref{fig-SSGC-loss-trace}, and \ref{fig-DGC-loss-trace}) for the three models with different values of $K$. Fig.~\ref{fig-SGC-loss-trace} and \ref{fig-SSGC-loss-trace} reflect that the decrease in the gradients' magnitudes will lead to the ineffective training of the models (thus leading to the over-smoothing phenomenon in Tab.~\ref{Cora results}, \ref{Cora-ml results}, \ref{Citeseer results} and \ref{Pubmed results}). For DGC, it is observed that the density plots of gradients are almost the same when $K>4$. The primary reason is that a small value of $T$ was chosen to achieve better classification performance (Sec.~E in the supplement provide the vanishing gradient issue when $T$ increases).  
This demonstrates that DGC can mitigate the vanishing gradient problem by properly choosing the parameter at the cost of sacrificing the performance when $K$ is small (Tab.~\ref{Cora results}, \ref{Cora-ml results}, \ref{Citeseer results} and \ref{Pubmed results}). Nevertheless, we show that our gradient-free method could still slightly improve DGC in some cases and improve the training efficiency.


%% file: sections/method.tex
\section{Gradient-free GCNs}
In this section, we elaborate on how to implement the \textbf{gradient-free} framework with the three linearized GCNs (SGC, SSGC and DGC) as its graph filters for the semi-supervised node classification task. 

\subsection{Graph Filters}
Theoretically, we can split the framework into two modules: feature extraction and kernelization. In the feature extraction module, we obtain different graph filters based on the graph's topological and semantic information:
\begin{equation}
    \label{fextra}
    \mathcal{F}_{G}=\begin{cases}
                                        \hat{\mathbf{A}}^{K}\mathbf{X}, & \mbox{for SGC,} 
                                        \\
                                        \frac{1}{K}\sum_{k=1}^{K}((1-\tau)\hat{\mathbf{A}}^{k}\mathbf{X}+\tau\mathbf{X}), & \mbox{for SSGC,}\\
                                        \left((1-\frac{T}{K})\mathbf{I}+\frac{T}{K}\hat{\mathbf{A}}\right)^{K}\mathbf{X}, & \mbox{for DGC}.
                                        \end{cases}
\end{equation}
After that, we can generalize and formulate the linearized GCNs as:
\begin{equation}
    \label{general-sgc}
    \widehat{\mathbf{Y}}=\text{softmax}(\mathcal{F}_{G}\mathbf{W}),
\end{equation}
where $\mathbf{W}$ is the single weight matrix to be trained.

\subsection{Kernelization}
After obtaining the corresponding graph filters, we endeavor to craft a gradient-free GCN model to overcome the previously mentioned vanishing gradient problem. Especially, the original objective function of the GCN model is the negative log-likelihood (NLL) loss, i.e.,
$\mathcal{L}_{NLL}=-\sum_{i=1}^{N}\sum_{c=1}^{C}y_{ic}\log\hat{y}_{ic}$.
Unfortunately, it has been proven that the NLL loss does not have its closed-form solution \cite{notcloseform}. Hence, a common way to optimize this objective is to utilize the vanilla gradient descent method like Adam~\cite{adam}, which highly relies on gradient computation. To avoid the vanishing gradient problem, it is natural to design a \textbf{gradient-free} method to optimize the model weights $\mathbf{W}$. To this end, we elaborately replace the NLL loss with the mean square loss augmented with the L2 penalty, i.e.,
\begin{equation}
\begin{split}
\label{eqn-OLS}
\mathcal{L}_{MSE}&\overset{\Delta}{=}\sum_{i=1}^{N}\sum_{c=1}^{C}\frac{1}{2}(y_{ic}-[\Phi(\mathcal{F}_{G})\mathbf{W}]_{ic})^{2}+\frac{\xi}{2} \Tr(\mathbf{W}^{\top}\mathbf{W})\\
&=\frac{1}{2}\Tr((\mathbf{Y}-\Phi(\mathcal{F}_{G})\mathbf{W})^{\top}(\mathbf{Y}-\Phi(\mathcal{F}_{G})\mathbf{W}))\\
 &\quad +\frac{\xi}{2} \Tr(\mathbf{W}^{\top}\mathbf{W}),
\end{split}
\end{equation}
where $\Phi(\mathcal{F}_{G})$ maps the graph filters to a new feature space, $\xi$ is the L2 regularizer to control the sparsity of the model. Henceforth, we attempt to find a closed-form solution $\mathbf{W}^*$ to optimize Eqn.~\eqref{eqn-OLS} without gradient computation. It is worth noting that in Eqn.~\eqref{eqn-OLS} the true label $\mathbf{Y}\in\mathbbm{R}^{N\times C}$ is a one-hot encoding matrix. That is, we transform the traditional multi-class classification problem to a \textbf{multiple multivariate regression} problem \cite{multiclassLR} by substituting our objective from the NLL loss to Eqn.~\eqref{eqn-OLS}, and we aim to find $\mathbf{W}^*$ which can push each entry of the prediction, i.e., $[\Phi(\mathcal{F}_{G})\mathbf{W}^*]_{ic}$ close to the binary entry $\mathbf{Y}_{ic}$. However, it is natural that $\Phi(\cdot)$ is unknown to us, we instead resort to the dual form of Eqn.~\eqref{eqn-OLS} and kernelize the inputs $\mathcal{F}_{G}$ to prevent the usage of $\Phi(\cdot)$. To tackle this issue, we reformulate Eqn.~\eqref{eqn-OLS} to:
\begin{equation}
\begin{split}
\label{eqn-dual-OLS}
&\mathcal{L}_{MSE}=\frac{1}{2}\Tr(\mathbf{Z}^{\top}\mathbf{Z})+\frac{\xi}{2} \Tr(\mathbf{W}^{\top}\mathbf{W}), \\
& \ \ \text{s.t.} \ \mathbf{Z}=\Phi(\mathcal{F}_{G})\mathbf{W}-\mathbf{Y}.
\end{split}
\end{equation}
We then obtain its Lagrange function:
\begin{equation}
\begin{split}
\label{eqn-Lagrange}
\mathcal{L}_{\bm{\Lambda}}&=\frac{1}{2}\Tr(\mathbf{Z}^{\top}\mathbf{Z})+\frac{\xi}{2} \Tr(\mathbf{W}^{\top}\mathbf{W})\\
&+\Tr(\bm{\Lambda}^{\top}(\mathbf{Z}-\Phi(\mathcal{F}_{G})\mathbf{W}+\mathbf{Y})),
\end{split}
\end{equation}
where $\bm{\Lambda}\in\mathbbm{R}^{N\times C}$ is the Lagrange multiplier. We then zero out the gradient $\frac{\partial\mathcal{L}_{\bm{\Lambda}}}{\partial\mathbf{Z}}$ and $\frac{\partial\mathcal{L}_{\bm{\Lambda}}}{\partial\mathbf{W}}$ and get:
\begin{equation}
\begin{split}
\label{eqn-dual-OLS-solution}
\mathbf{Z}^{*}=-\bm{\Lambda}, \quad \mathbf{W}^*=\frac{1}{\xi}\Phi(\mathcal{F}_{G})^{\top}\bm{\Lambda}.
\end{split}
\end{equation}
Next, by substituting $\mathbf{Z}$ and $\mathbf{W}$ in Eqn.~\eqref{eqn-Lagrange} with Eqn.~\eqref{eqn-dual-OLS-solution} we obtain the dual form of Eqn.~\eqref{eqn-OLS}, i.e.,
\begin{equation}
\begin{split}
\label{eqn-alpha-OLS}
\mathcal{L}_{\bm{\Lambda}}&=-\frac{1}{2}\Tr(\bm{\Lambda}^{\top}\bm{\Lambda})-\frac{1}{2\xi}\Tr(\bm{\Lambda}^{\top}\mathbf{M}\bm{\Lambda})+\Tr(\bm{\Lambda}^{\top}\mathbf{Y}),
\end{split}
\end{equation}
where $\mathbf{M}$ is the kernel matrix based on the inputs $\mathcal{F}_{G}$, i.e., $\mathbf{M}_{uv}=m(\mathcal{F}_{Gu},\mathcal{F}_{Gv})$, $m(\cdot)$ is the kernel function. 

\subsection{Closed-form Solution}
Given the dual form loss function Eqn.~\eqref{eqn-alpha-OLS}, we aim to find an optimal $\bm{\Lambda}^{*}$ to optimize this objective in a gradient-free manner. To tackle this issue, we compute the gradient of $\frac{\partial\mathcal{L}_{\bm{\Lambda}}}{\partial\bm{\Lambda}}$ and zeroize it to explicitly represent $\bm{\Lambda}^*$ with the kernel matrix:
\begin{equation}
\begin{split}
\label{eqn-alpha-closeform}
\bm{\Lambda}^*&=(\frac{1}{\xi}\mathbf{M}+\mathbf{I})^{-1}\mathbf{Y}\overset{\Delta}{=}(\lambda\mathbf{M}+\mathbf{I})^{-1}\mathbf{Y},
\end{split}
\end{equation}
After that, we can predict the node label in the testing set based on the optimal solution $\bm{\Lambda}$:
\begin{equation}
\begin{split}
\label{eqn-alpha-pred}
\bm{y}^{*}=\Phi(\mathcal{F}_{G})\mathbf{W}^{*}=\lambda\cdot m(\mathcal{F}_{G}^{*},\mathcal{F}_{G}^{train})\bm{\Lambda}^{*},
\end{split}
\end{equation}
where $\mathcal{F}_{G}^{*}$ is the graph filter of a new node computed based on Eqn.~\eqref{fextra}, $\mathcal{F}_{G}^{train}$ is the graph filter of training set, $\bm{y}^{*}\in\mathbbm{R}^{1\times C}$ is prediction of the new node. For evaluation, we pick out the index of the largest entry of $\bm{y}^*$ as the predicted label for the new node. In addition to that, if it is a binary classification problem,  we can still use Eqn.~\eqref{eqn-alpha-pred} for prediction. The only difference between binary classification and multi-class classification is that the binary classification does not utilize the one-hot encoding of the true label $\mathbf{Y}$. It is worth noting that if the kernel function $m(\cdot)$ is a linear kernel, we have the primal version of the closed-form solution to the weight $\mathbf{W}$ in Eqn.~\ref{eqn-OLS}, i.e.,
$\mathbf{W}^*=(\mathcal{F}_{G}^{\top}\mathcal{F}_{G}+\xi\mathbf{I})^{-1}\mathcal{F}_{G}^{\top}\mathbf{Y}$.

\subsection{Scale Invariance of the Closed-form Solution}
\label{sec-scale-invariance}
In the primal form, the optimization problem \eqref{eqn-OLS} has following closed-form solution,
\begin{equation}
\mathbf{W}^* = \left(\Phi(\mathcal{F}_{G})^\top \Phi(\mathcal{F}_{G}) + \xi I\right)^{-1}\Phi(\mathcal{F}_{G})^\top \mathbf{Y}.
\end{equation}
The optimal solution $\mathbf{W}^*$ results in an estimation $\widehat{\mathbf{Y}}^*$,
\begin{equation}
\begin{split}
    \widehat{\mathbf{Y}}^*
        & = \Phi(\mathcal{F}_{G})\mathbf{W}^* \\
        & = \Phi(\mathcal{F}_{G})\left(\Phi(\mathcal{F}_{G})^\top \Phi(\mathcal{F}_{G}) + \xi I\right)^{-1}\Phi(\mathcal{F}_{G})^\top \mathbf{Y}.
\end{split}
\end{equation}
We consider $\widehat{\mathbf{Y}}^*$ as a function of $\Phi(\mathcal{F}_{G})$ and $\xi$,
\begin{equation}
    \widehat{\mathbf{Y}}^* = \widehat{\mathbf{Y}}^*(\Phi(\mathcal{F}_{G}), \xi).
\end{equation}
If we scale $\Phi(\mathcal{F}_{G})$ by a constant $\beta$, and scale $\xi$ by $\beta^2$, the estimation $\widehat{\mathbf{Y}}^*$ is invariant, i.e.,
\begin{equation}
    \widehat{\mathbf{Y}}^* (\Phi(\mathcal{F}_{G}), \xi) = \widehat{\mathbf{Y}}^* (\beta \Phi(\mathcal{F}_{G}), \beta^2 \xi).
\end{equation}
In other words, the estimator resulting from the optimization problem \eqref{eqn-OLS} is invariant to the scale of $\Phi(\mathcal{F}_{G})$, as long as hyper-parameter $\xi$ is properly scaled correspondingly. 

%% file: sections/exp.tex
\section{Experiments}
\subsection{Dataset Description}
We consider the node classification task over four commonly used citation networks: Cora, Cora-ML, Citeseer, and PubMed \cite{Datasets}. The nodes in these datasets represent published papers, and the edges are citations among papers. More details about the datasets are presented in Sec.~A of the supplement. 

\subsection{Baselines}
To demonstrate the performance of the proposed model, we compare our models with eight strong baselines, including GCN \cite{GCN}, Graph Attention Network (GAT)~\cite{GAT}, Jumping Knowledge Networks(JKNet)~\cite{JKNet}, APPNP~\cite{APPNP}, Graph Convolutional Network via Initial residual and Identity mapping (GCNII)~\cite{GCNII}, SGC~\cite{SGC}, SSGC~\cite{SSGC} and DGC~\cite{DGC}. It is worth noting that GCN, GAT, JKNet, APPNP, GCNII are nonlinear deep learning models; SGC, SSGC and DGC are linearized models.

\subsection{Model Setting}
For a fair comparison, each method is run on an Ubuntu 20.04 system with 16 Intel(R) Xeon(R) Platinum 8369B CPU @ 2.90GHz and a GeForce RTX 3090. The computer codes and the parameters used for the GCN, GAT, SGC and SSGC are available from the authors' public links. We choose the default hyperparameter settings for SSGC and DGC, i.e., we set the strength of identity mapping $\tau=0.05$ for SSGC and the terminal time $T=5.27$ for DGC. For all the models to be trained via gradient descent, we utilize the Adam~\cite{adam} with the learning rate $0.01$ and epochs equal to $200$. For each of the datasets, we split the dataset into a training, validation and testing set with a proportion equal to $1:1:8$. We run all the models $20$ times with different seeds and report the mean testing accuracy for evaluation. Without loss of generality, we choose the kernel function $m(\cdot)$ as the linear kernel for a fair comparison. We term our methods with three different graph filters as: gfSGC, gfSSGC and gfDGC respectively. 

\subsection{Experiment Results}
\subsubsection{Overall Node Classification Performance}
Our gradient-free method results in three refined linearized GCN models: gfSGC, gfSSGC, and gfDGC. These models allow one to choose larger values of $K$ to incorporate more information during the message-passing process. 
To validate the effectiveness of our models, we compare them with eight baselines (including both linearized and nonlinear models) on the node classification task. We note that, for all models, we report their best performance among different values of $K$, where $K\in\{2,4,8,16,32,64,128\}$. The results are summarized in Tab.~\ref{All results}.

It is observed that our gradient-free methods gfSGC, gfSSGC, and gfDGC achieve the best accuracies among the strong baselines in almost all cases, often with a clear improvement over other GCNs. For instance, the best accuracies of our models on Cora, Cora-ML, Citeseer, and PubMed are $86.5\%$, $87.5\%$, $77.4\%$, and $87.0\%$ respectively. 
One exceptional case is on Pubmed, where the best model APPNP is a nonlinear GNN model. However, we note that linearized GCNs are much more efficient than nonlinear models in terms of training.
It is noteworthy that the gradient-free methods are always better than the baseline models, i.e., SGC, SSGC and DGC. 
This observation directly demonstrates the advantages and effectiveness of the gradient-free framework.  
 
\begin{table}[htbp]
    \caption{Average test accuracy (\%$\pm$ standard error) over $20$ runs on four node classification datasets. The best and second-best results are highlighted in bold and underlined.}
    \centering
    \label{All results}
    \resizebox{0.9\columnwidth}{!}{%
    \begin{tabular}{l|rrrr}
        \toprule[1.pt]
        Datasets  & Cora & Cora-ML & Citeseer & Pubmed\\
        \hline
        GCN  & $83.6\pm{0.6}$ & $85.7\pm{0.4}$ & $71.6\pm{0.4}$ & $85.0\pm{0.2}$ \\
        GAT  & $83.7\pm{0.3}$ & $84.6\pm{0.5}$ & $73.1\pm{2.3}$ & $83.7\pm{0.5}$ \\
        JKNet  & $85.0\pm{0.5}$ & $86.4\pm{0.3}$ & $72.4\pm{0.5}$ & $86.3\pm{0.1}$ \\
        APPNP  & $85.5\pm{0.5}$ & $86.8\pm{0.7}$ & $71.8\pm{0.7}$ & $\bm{87.1\pm{0.3}}$ \\
        GCNII  & $79.2\pm{0.4}$ & $82.7\pm{1.1}$ & $70.4\pm{0.4}$ & $86.9\pm{0.3}$ \\
        \hline
        SGC  & $85.3\pm{0.06}$ & $86.1\pm{0.03}$ & $74.1\pm{0.1}$ & $82.7\pm{0.03}$ \\
        SSGC  & $85.9\pm{0.07}$ & \bm{$87.5\pm{0.05}$} & $75.2\pm{0.02}$ & $82.7\pm{0.03}$ \\
        DGC  & $85.2\pm{0.03}$ & $86.2\pm{0.04}$ & $73.8\pm{0.06}$ & $81.2\pm{0.03}$ \\
        \hline
        gfSGC  & \underline{$86.4$} & \underline{$87.3$} & $\bm{77.4}$ & $86.7$\\
        gfSSGC  & $\bm{86.5}$ & $\bm{87.5}$ & $\bm{77.4}$ & \underline{$87.0$}\\
        gfDGC  & $\underline{86.4}$ & $86.4$ & $\underline{76.8}$ & $84.7$\\
        \bottomrule[1.pt]
    \end{tabular}
    }
\end{table}

\begin{table*}[htbp]
    \centering
    \caption{Average test accuracy (\%$\pm$ standard error) over $20$ runs on Cora dataset. \protect\\ The best and second-best results are highlighted in bold and underlined.}
    \label{Cora results}
    \resizebox{1.8\columnwidth}{!}{
    \begin{tabular}{l|rrrrrrr}
        \toprule[1pt]
        Model  & $K=2$ & $K=4$ & $K=8$ & $K=16$ & $K=32$ & $K=64$ & $K=128$\\
        \hline
        GCN  & $83.6\pm{0.6}$ & $81.2\pm{0.9}$ & $69.4\pm{3.7}$ & $39.3\pm{7.6}$ & $30.2\pm{3.1}$ & $29.2\pm{0.02}$ & $29.2\pm{0.02}$ \\
        GAT  & $83.7\pm{0.3}$ & $82.4\pm{0.9}$ & $81.8\pm{1.0}$ & $56.9\pm{16.6}$ & $33.0\pm{8.9}$ & $29.2\pm{9.3}$ & $11.5\pm{0}$\\
        GCNII  & $79.1\pm{1.0}$ & $77.8\pm{1.2}$ & $78.9\pm{0.4}$ & $79.0\pm{0.5}$ & $79.2\pm{0.4}$ & $79.1\pm{0.5}$ & $79.1\pm{0.3}$\\
        SGC  & $84.9\pm{0.1}$ & $85.3\pm{0.06}$ & $84.5\pm{0.07}$ & $83.2\pm{0.6}$ & $80.7\pm{0.8}$ & $73.2\pm{0.2}$ & $61.3\pm{0.3}$\\
        SSGC  & $83.5\pm{0.1}$ & $85.5\pm{0.1}$ & $85.8\pm{0.08}$ & \underline{$85.9\pm{0.07}$} & $85.2\pm{0.08}$ & $84.7\pm{0.05}$ & $84.4\pm{0.09}$\\
        DGC  & $45.0\pm{0.3}$ & $84.6\pm{0.06}$ & $85.2\pm{0.07}$ & $85.2\pm{0.04}$ & $85.2\pm{0.04}$ & $85.2\pm{0.03}$ & $85.2\pm{0.05}$\\
        \hline
        gfSGC  & $\bm{86.4}$ & \underline{$86.4$} & $84.9$ & $83.7$ & $82.0$ & $80.2$ & $77.6$\\
        gfSSGC & \underline{$85.4$} & \bm{$86.5$} & \bm{$86.5$} & \bm{$86.4$} & \underline{$86.3$} & \underline{$86.2$} & \underline{$86.0$}\\
        gfDGC  & $39.1$ & $84.9$ & \underline{$86.4$} & \bm{$86.4$} & \bm{$86.4$} & \bm{$86.3$} & \bm{$86.2$}\\
        \bottomrule[1.pt]
    \end{tabular}
    }
\end{table*}

\begin{table*}[htbp]
    \centering
    \caption{Average test accuracy (\%$\pm$ standard error) over $20$ runs on Cora-ML dataset. \protect\\ The best and second-best results are highlighted in bold and underlined.}
    \label{Cora-ml results}
    \resizebox{1.8\columnwidth}{!}{
    \begin{tabular}{l|rrrrrrr}
        \toprule[1.pt]
        Model  & $K=2$ & $K=4$ & $K=8$ & $K=16$ & $K=32$ & $K=64$ & $K=128$\\
        \hline
        GCN  & $85.7\pm{0.4}$ & $82.3\pm{1.7}$ & $66.3\pm{10.5}$ & $55.8\pm{6.7}$ & $38.0\pm{7.1}$ & $32.9\pm{6.1}$ & $27.8\pm{0.4}$\\
        GAT  & $84.6\pm{0.5}$ & $83.1\pm{1.1}$ & $81.5\pm{5.3}$ & $54.9\pm{17.7}$ & $27.4\pm{3.9}$ & $17.4\pm{7.8}$ & $12.4\pm{0}$\\
        GCNII  & $82.7\pm{1.1}$ & $82.7\pm{1.2}$ & $79.2\pm{1.2}$ & $79.5\pm{1.3}$ & $80.6\pm{1.4}$ & $78.8\pm{1.5}$ & $79.1\pm{1.1}$\\
        SGC  & $86.0\pm{0.05}$ & $86.1\pm{0.03}$ & $84.6\pm{0.03}$ & $81.2\pm{0.06}$ & $73.5\pm{0.02}$ & $61.3\pm{0.02}$ & $38.1\pm{0.1}$\\
        SSGC  & $85.9\pm{0.08}$ & \underline{$86.6\pm{0.06}$} & \bm{$87.5\pm{0.05}$} & \underline{$86.6\pm{0.03}$} & $85.4\pm{0.07}$ & $83.0\pm{0.04}$ & $77.0\pm{0.05}$\\
        DGC  & $50.6\pm{0.07}$ & $85.4\pm{0.06}$ & $85.8\pm{0.04}$ & $86.1\pm{0.04}$ & $86.1\pm{0.06}$ & \underline{$86.2\pm{0.04}$} & \underline{$86.2\pm{0.03}$}\\
        \hline
        gfSGC  & \bm{$87.3$} & $86.4$ & $84.7$ & $84.3$ & $83.0$ & $80.5$ & $72.8$\\
        gfSSGC & \underline{$86.2$} & \bm{$87.0$} & \bm{$87.5$} & \bm{$86.9$} & \bm{$86.3$} & \bm{$86.3$} & \bm{$86.4$}\\
        gfDGC  & $41.4$ & $85.3$ & \underline{$86.1$} & $86.1$ & \underline{$86.2$} & \bm{$86.3$} & \bm{$86.4$}\\
        \bottomrule[1.pt]
    \end{tabular}
    }
\end{table*}

\begin{table*}[htbp]
    \centering
    \caption{Average test accuracy (\%$\pm$ standard error) over $20$ runs on Citeseer dataset. \protect\\ The best and second-best results are highlighted in bold and underlined.}
    \label{Citeseer results}
    \resizebox{1.8\columnwidth}{!}{
    \begin{tabular}{l|rrrrrrr}
        \toprule[1.pt]
        Model  & $K=2$ & $K=4$ & $K=8$ & $K=16$ & $K=32$ & $K=64$ & $K=128$\\
        \hline
        GCN  & $71.6\pm{0.4}$ & $69.2\pm{0.9}$ & $62.5\pm{3.4}$ & $54.3\pm{8.1}$ & $38.0\pm{13.2}$ & $26.6\pm{2.1}$ & $26.3\pm{2.7}$\\
        GAT  & $73.1\pm{2.3}$ & $70.4\pm{4.8}$ & $63.7\pm{13.4}$ & $57.4\pm{8.2}$ & $30.6\pm{14.5}$ & $8.3\pm{6.7}$ & $5.5\pm{0}$\\
        GCNII  & $67.7\pm{1.0}$ & $67.9\pm{0.9}$ & $68.9\pm{0.4}$ & $70.3\pm{0.6}$ & $70.4\pm{0.4}$ & $70.4\pm{0.4}$ & $70.3\pm{0.6}$\\
        SGC  & $74.1\pm{0.1}$ & $73.2\pm{0.06}$ & $72.7\pm{0.09}$ & $73.3\pm{0.07}$ & $72.0\pm{0.06}$ & $71.1\pm{0.02}$ & $72.0\pm{0.03}$\\
        SSGC  & \underline{$74.7\pm{0.1}$} & $74.7\pm{0.1}$ & $74.5\pm{0.09}$ & $75.2\pm{0.06}$ & $75.2\pm{0.08}$ & $75.2\pm{0.02}$ & $74.2\pm{0.05}$\\
        DGC  & $44.0\pm{0.3}$ & $73.8\pm{0.08}$ & $73.7\pm{0.07}$ & $73.8\pm{0.06}$ & $73.7\pm{0.05}$ & $73.7\pm{0.06}$ & $73.7\pm{0.06}$\\
        \hline
        gfSGC  & \bm{$77.4$} & \underline{$76.7$} & $75.7$ & $75.1$ & $74.4$ & $73.1$ & $72.5$\\
        gfSSGC & \bm{$77.4$} & \bm{$77.4$} & \bm{$77.3$} & \underline{$76.4$} & \underline{$76.1$} & \underline{$75.7$} & \underline{$75.4$}\\
        gfDGC  & $46.0$ & $76.5$ & \underline{$76.7$} & \bm{$76.8$} & \bm{$76.7$} & \bm{$76.7$} & \bm{$76.6$}\\
        \bottomrule[1.pt]
    \end{tabular}
    }
\end{table*}

\begin{table*}[htbp]
    \centering
    \caption{Average test accuracy (\%$\pm$ standard error) over $20$ runs on PubMed dataset. \protect\\ The best and second-best results are highlighted in bold and underlined.}
    \label{Pubmed results}
    \resizebox{1.8\columnwidth}{!}{
    \begin{tabular}{l|rrrrrrr}
        \toprule[1.pt]
        Model  & $K=2$ & $K=4$ & $K=8$ & $K=16$ & $K=32$ & $K=64$ & $K=128$\\
        \hline
        GCN  & $85.0\pm{0.2}$ & $82.6\pm{0.4}$ & $72.4\pm{15.6}$ & $56.2\pm{15.8}$ & $44.0\pm{3.4}$ & $44.0\pm{2.8}$ & $45.4\pm{4.9}$\\
        GAT  & $83.7\pm{0.5}$ & $82.0\pm{0.9}$ & $81.0\pm{4.7}$ & $61.8\pm{19.1}$ & $41.4\pm{5.6}$ & $32.3\pm{9.3}$ & $21.7\pm{4.2}$\\
        GCNII  & \underline{$86.9\pm{0.3}$} & \underline{$85.9\pm{0.4}$} & \underline{$85.6\pm{0.4}$} & \underline{$85.3\pm{0.3}$} & \underline{$84.7\pm{0.3}$} & $84.4\pm{0.2}$ & \underline{$85.3\pm{0.2}$}\\
        SGC  & $82.7\pm{0.03}$ & $81.2\pm{0.04}$ & $79.9\pm{0.04}$ & $78.1\pm{0.04}$ & $73.0\pm{0.1}$ & $64.1\pm{0.2}$ & $58.1\pm{0.1}$\\
        SSGC  & $82.7\pm{0.03}$ & $82.7\pm{0.03}$ & $82.0\pm{0.02}$ & $81.3\pm{0.03}$ & $80.2\pm{0.03}$ & $77.9\pm{0.07}$ & $72.9\pm{0.1}$\\
        DGC  & $68.8\pm{0.03}$ & $80.5\pm{0.04}$ & $81.0\pm{0.05}$ & $81.1\pm{0.03}$ & $81.1\pm{0.03}$ & $81.2\pm{0.03}$ & $81.2\pm{0.03}$\\
        \hline
        gfSGC  & $86.7$ & $84.7$ & $83.3$ & $82.1$ & $81.0$ & $80.0$ & $77.6$\\
        gfSSGC & \bm{$87.0$} & \bm{$87.0$} & \bm{$86.2$} & \bm{$85.8$} & \bm{$85.4$} & \bm{$85.3$} & \bm{$85.4$}\\
        gfDGC  & $52.1$ & $82.6$ & $84.6$ & $84.7$ & $84.7$ & \underline{$84.6$} & $84.6$\\
        \bottomrule[1.pt]
    \end{tabular}
    }
\end{table*}

\subsubsection{Mitigating Over-smoothing}
We further provide the node classification accuracies with varying depth $K\in\{2, 4, 8, 16, 32, 64, 128\}$ in Tab.~\ref{Cora results}, \ref{Cora-ml results}, \ref{Citeseer results} and  \ref{Pubmed results} to demonstrate the over-smoothing issue  as $K$ increases. Specifically, it is observed that the accuracies of vanilla baselines such as  GCN, GAT, and SGC decrease sharply with the increases in depths. For instance, the accuracies of GCN and GAT on Cora decrease to about $30\%$ when $K>16$. In some extreme cases, such as GAT on Citeseer when $K=64$ and $128$ the accuracies are below $10\%$. 
In general, those baselines (e.g., SSGC and DGC) specifically designed to overcome over-smoothing perform much more stable as $K$ increases.  In particular, the accuracies of DGC are almost the same when $K>4$. This phenomenon is consistent with the empirical results in Fig.~\ref{fig-DGC-initial-gradients} and \ref{fig-DGC-loss-trace}. That is, DGC mitigates the vanishing gradient issue to prevent over-smoothing.

In contrast to the baselines, our gradient-free models, i.e., gfSGC, gfSSGC and gfDGC maintain almost invariant performance as $K$ increases. For example, on Cora, the accuracy of gfSGC decreases from $86.4\%$ to $77.6\%$ when depth $K$ increases from $2$ to $128$, while SGC greatly decreases from $84.9\%$ to $61.3\%$. Even for DGC where over-smoothing is not quite obvious, our counterpart gfDGC can achieve comparable or slightly better performance. Similar cases can be found for other models on other datasets. An interesting observation is that GCNII achieves the second-best performance on PubMed under all layer numbers. One possible reason is that large-scale dataset such as PubMed prefers the complex architecture of GCNII. 

\subsubsection{Gradient-free v.s. Linear Models}
Our gradient-free models gfSGC, gfSSGC and gfDGC can be regarded as the refinement of their counterparts SGC, SSGC, and DGC, respectively.  To validate the effectiveness of the gradient-free method, it is necessary to directly compare gfSGC, gfSSGC, and gfDGC with the linear models with the corresponding filters. 

We note that SGC and gfSGC share the same feature extraction process and the extracted graph filter of SGC is not specially designed to tackle over-smoothing. Notably, the performance of SGC drastically decreases around $27.7\%$ and $55.7\%$ for Cora and Cora-ML while gfSGC slightly decreases around $10.2\%$ and $16.6\%$. This fact directly demonstrates the effectiveness of the gradient-free method in mitigating vanishing gradients and further over-smoothing.
As for SSGC, although it crafts the graph filter to balance the global and local context of each node to mitigate over-smoothing, augmenting SSGC with the gradient-free method can further enhance their performances and robustness with varying depths. 
For example, SSGC drastically decreases around $10.4\%$ and $11.9\%$ for Cora-ML and PubMed while gfSSGC slightly declines around $-0.7\%$ and $1.8\%$. These results are consistent with the empirical results in Fig.~\ref{fig-SSGC-initial-gradients} and \ref{fig-SSGC-loss-trace}. 
Since DGC itself can mitigate the vanishing gradient issue quite well, gfDGC just slightly improves the performance of DGC. In addition, we also observe that gfSSGC (i.e., gradient-free built upon a different graph filter) outperforms DGC. For example, the gaps between the best performance of gfSSGC and DGC for these four datasets are $1.3\%$, $1.3\%$, $3.7\%$, and $5.8\%$, respectively. It demonstrates 
that the gradient-free method could potentially be a better way in tackling the vanishing gradient problem, thus leading to a better linearized GCN model.


\subsubsection{Ablation Study}
\begin{figure}[h]
	\centering
	\begin{subfigure}[b]{0.234\textwidth}
		\centering
		\includegraphics[width=\textwidth,height=2cm]{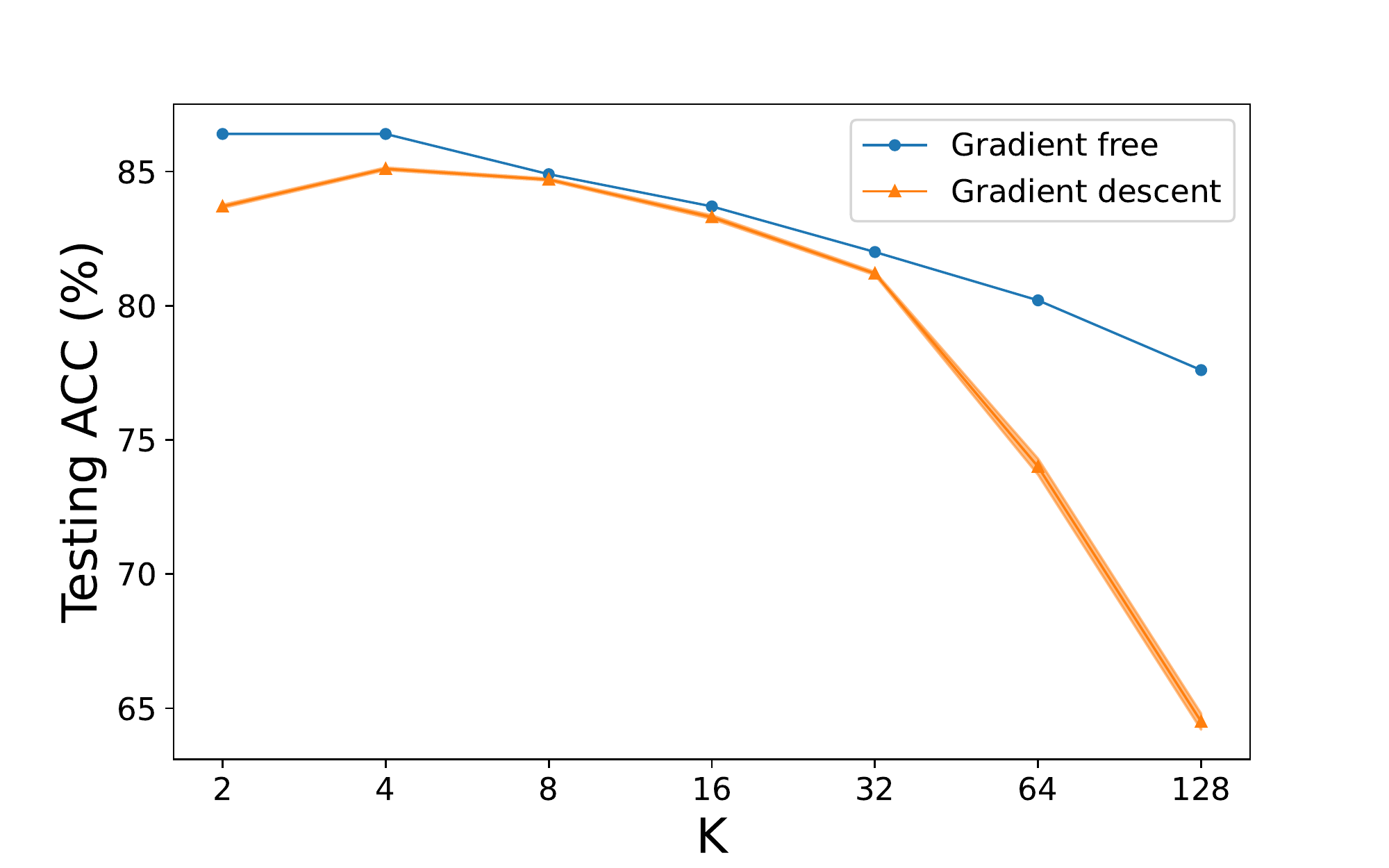}
		\caption{Cora}
        \label{fig-ablation-cora}
	\end{subfigure}
	\hfill
	\begin{subfigure}[b]{0.234\textwidth}
		\centering
		\includegraphics[width=\textwidth,height=2cm]{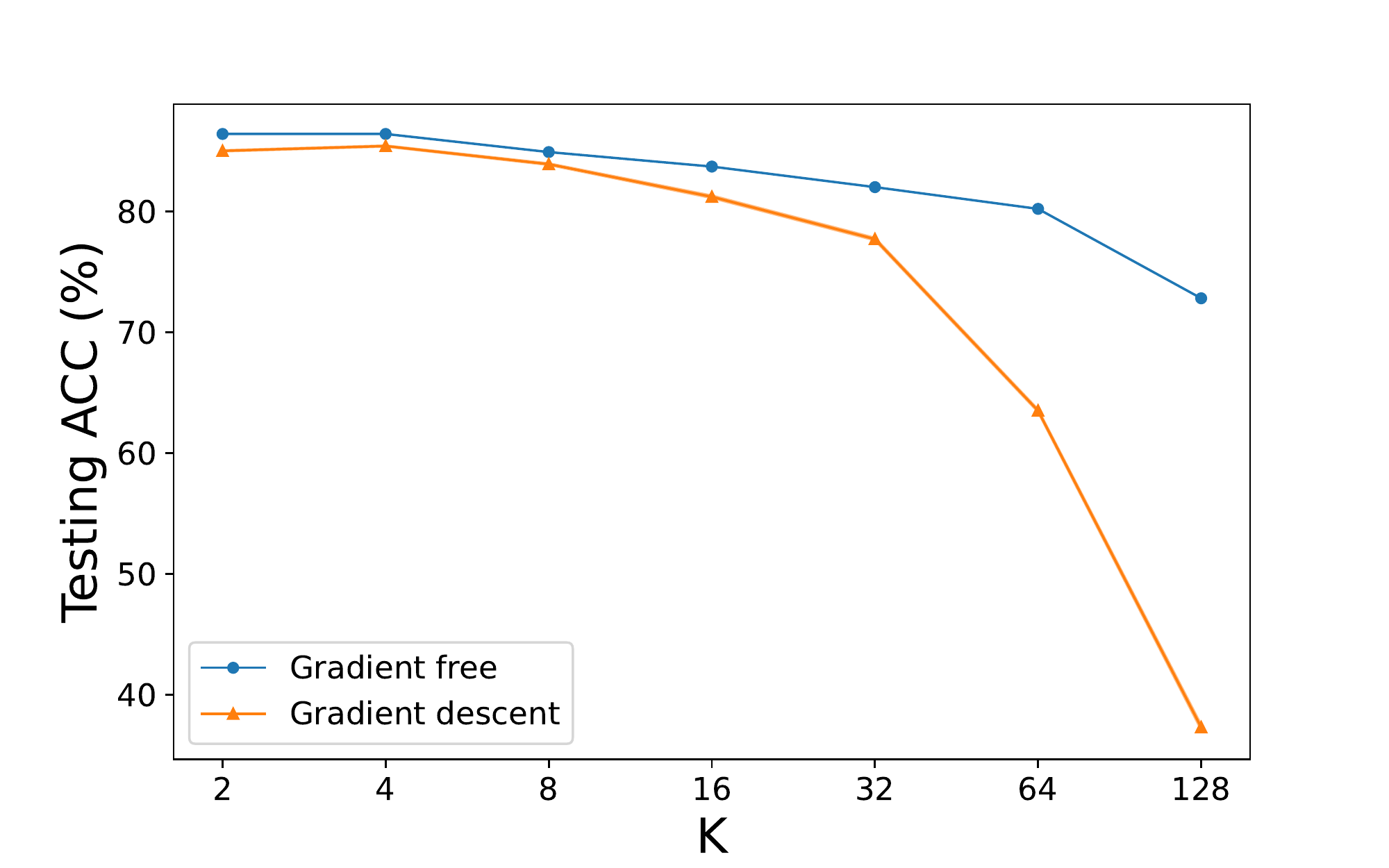}
		\caption{Cora-ML}
        \label{fig-ablation-cora-ml}
	\end{subfigure}
	\hfill
	\begin{subfigure}[b]{0.234\textwidth}
		\centering
		\includegraphics[width=\textwidth,height=2cm]{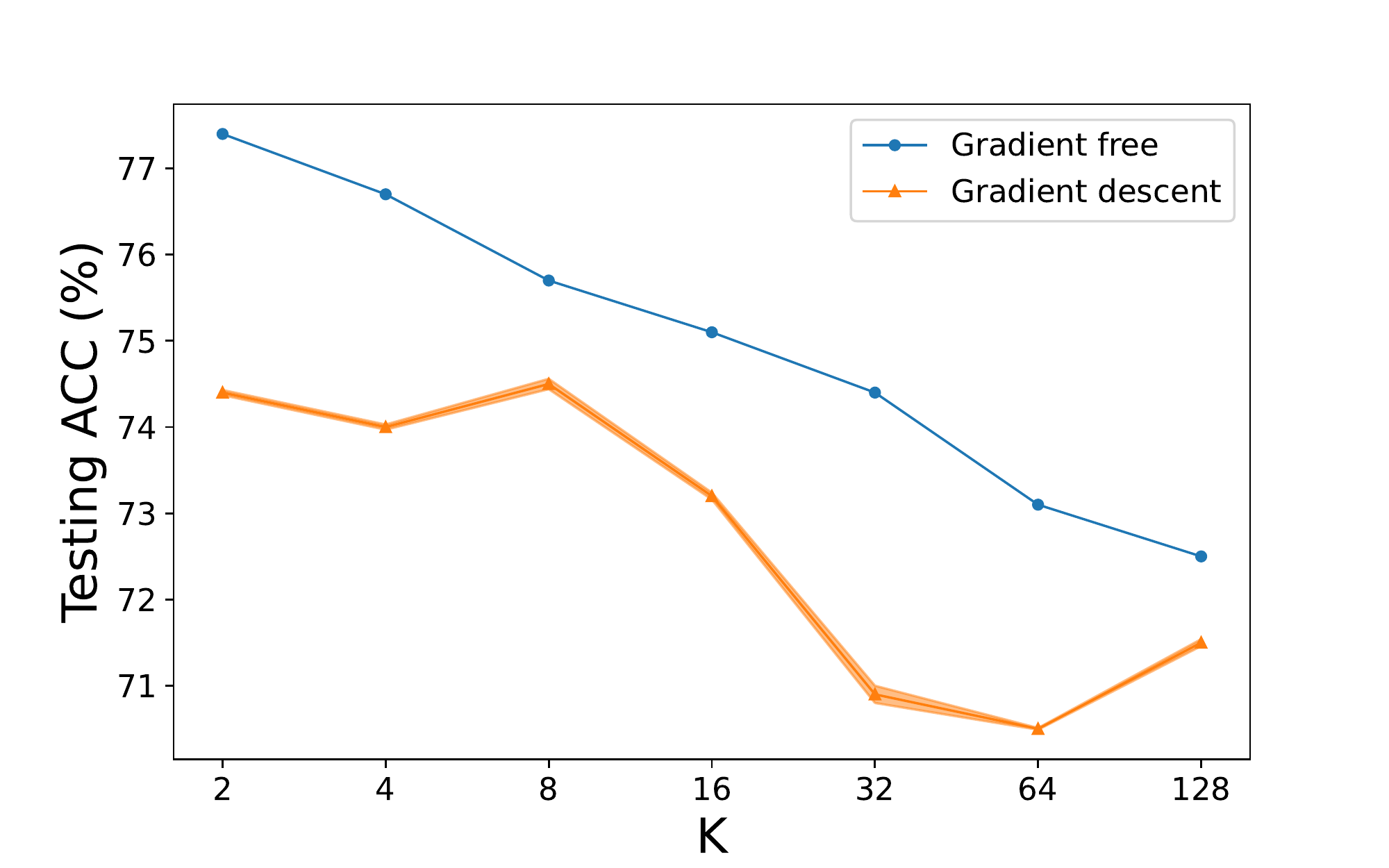}
		\caption{Citeseer}
        \label{fig-ablation-citeseer}
	\end{subfigure}
	\hfill
	\begin{subfigure}[b]{0.234\textwidth}
		\centering
		\includegraphics[width=\textwidth,height=2cm]{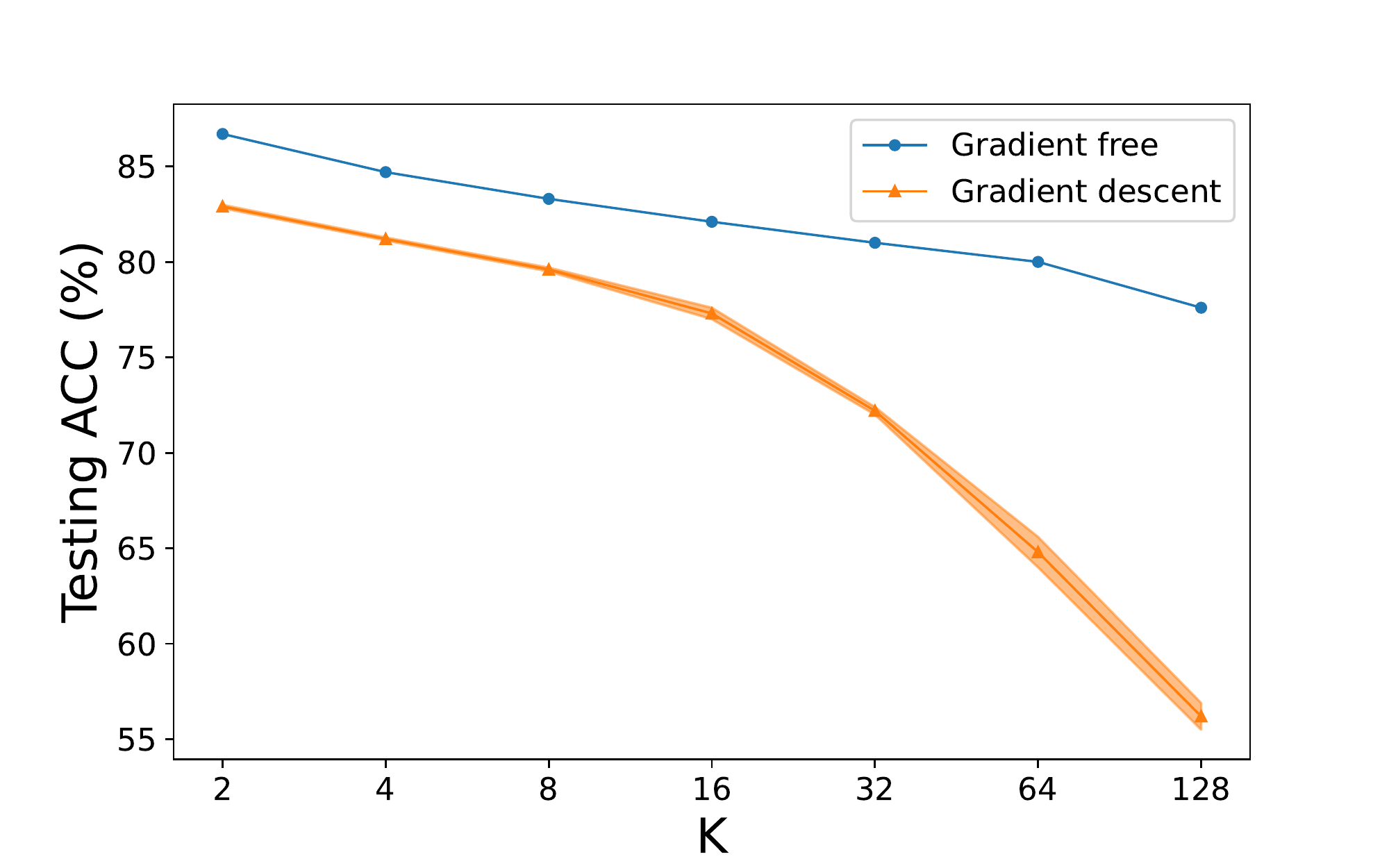}
		\caption{Pubmed}
        \label{fig-ablation-pubmed}
	\end{subfigure}
	\caption{Gradient descent v.s. gradient-free with SGC filter.}
	\label{fig-ablation-study}
\end{figure}
We conduct the ablation study to validate the necessity of the gradient-free method for the linearized GCNs. Specifically, we utilize the gradient descent to optimize $\Lambda$ in Eqn.~\eqref{eqn-alpha-closeform} and term it ``Gradient descent". The experiment results are shown in Fig.~\ref{fig-ablation-study}. For each case, we set the same values for the hyperparameter $\lambda$ for ``Gradient free" and ``Gradient descent" and set sufficient epoch for ``Gradient descent" to ensure convergence. Without loss of generality, We use the Cora dataset as an example. The experiment results demonstrate that ``Gradient free" outperforms the gradient descent method for all different depths especially when $K$ is large. On the other hand, for Cora, Cora-ML and PubMed, there is a trend that the gap between ``Gradient free" and ``Gradient descent" becomes larger with the decreasing of the depth $K$. This indicates that our method indeed mitigates the over-smoothing problem by preventing gradient computing during training.

\subsection{Other Benefits of Gradient-Free}
\subsubsection{Comparison of Stability}
The advantages of our methods also include stability. As shown in Tab.~\ref{Cora results}, \ref{Cora-ml results}, \ref{Citeseer results}, \ref{Pubmed results}, existing GCNs are unstable in some cases, which is reflected in high standard errors. For example, on Cora-ml, the standard errors of GCNII are more than $1.0$. Benefiting from the closed-form solution in Eqn.~\eqref{eqn-alpha-closeform}, the closed-form solution is a deterministic mapping with the inputs as kernel matrix, one-hot encoding label matrix and the hyperparameter $\lambda$ and is regardless of the initial values of the weight matrix. Hence, setting different random seeds will lead to the same classification results and the standard error is zero. 

\subsubsection{Comparison of Efficiency}
\begin{table}[h]
    \centering
    \caption{Training time cost (s) on PubMed dataset.}
    \resizebox{0.8\columnwidth}{!}{
    \label{Pubmed time cost}
    \begin{tabular}{l|rrrr}
        \toprule[1.pt]
        Model  & $K=2$ & $K=8$ & $K=32$ & $K=128$\\
        \hline
        GCN  & $1.67$ & $1.96$ & $6.58$ & $24.89$\\
        GCNII& $4.10$ & $6.30$ & $20.37$ & $77.57$\\
        SGC  & $0.24$ & $0.26$ & $0.30$ & $0.46$\\
        SSGC & $0.24$ & $0.26$ & $0.30$ & $0.47$\\
        DGC  & $0.25$ & $0.26$ & $0.30$ & $0.50$\\
        \hline
        gfSGC & $0.02$ & $0.02$ & $0.02$ & $0.02$\\
        gfSSGC& $0.02$ & $0.02$ & $0.02$ & $0.02$\\
        gfDGC & $0.02$ & $0.02$ & $0.02$ & $0.02$\\
        \bottomrule[1.pt]
    \end{tabular}
    }
\end{table}
In this section, we evaluate the training efficiency of our methods compared with other baselines. For gradient free methods, we report the time cost of closed-form solution, while for other baselines we report the training time. It is observed from Tab.~\ref{Pubmed time cost} that the linearized GCNs are much more efficient than the nonlinear models (GCN and GCNII). Comparing the gradient-free methods and linear models, it is demonstrated that utilizing the closed-form solution in a one-pipeline manner is more time-efficient than the vanilla gradient descent.



%% file: sections/conclusion.tex
\section{Conclusions}
In this paper, we propose a gradient-free training algorithm to provide significant improvements over the linearized GCN models through kernelization and closed-form solutions to the model weights. Sufficient experiments demonstrate that our methods outperform other typical baselines with less time cost. The intuition is that current linearized GCNs still perform not well when the model depth is large under some scenarios. The theoretical and empirical finding is that when the model depth is large, the gradients of the model weights tend to be minuscule -- ``vanishing gradient". We hope that this finding will provide new insights to the community to propose alternative views on tackling the ``over-smoothing" problem and boost the performance of the linearized GCNs. 